%% file: main.tex
\documentclass[10pt,twocolumn,letterpaper]{article}

\usepackage{cvpr}              

\input{preamble}

\usepackage{bm}
\usepackage{soul}
\usepackage{multirow}

\definecolor{cvprblue}{rgb}{0.21,0.49,0.74}
\usepackage[pagebackref,breaklinks,colorlinks,citecolor=cvprblue]{hyperref}

\def\paperID{11831} 
\def\confName{CVPR}
\def\confYear{2024}

\title{Noise Adaptor in Spiking Neural Networks}

\author{Chen Li\\
King’s College London\\
London, UK\\
{\tt\small chen.7.li@kcl.ac.uk}
\and
Bipin.Rajendran\\
King’s College London\\
London, UK\\
{\tt\small bipin.rajendran@kcl.ac.uk}
}

\begin{document}
\maketitle
\input{sec/0_abstract}    
\input{sec/1_intro}

\input{sec/2_formatting}

\input{sec/3_finalcopy}

\input{sec/4}

\input{sec/5}

\input{sec/7}
{
    \small
    \bibliographystyle{ieeenat_fullname}
    \bibliography{main}
}


\end{document}

%% file: preamble.tex
\usepackage[dvipsnames]{xcolor}


%% file: sec/0_abstract.tex
\begin{abstract}
Recent strides in low-latency spiking neural network (SNN) algorithms have drawn significant interest, particularly due to their event-driven computing nature and fast inference capability. One of the most efficient ways to construct a low-latency SNN is by converting a pre-trained, low-bit artificial neural network (ANN) into an SNN. However, this conversion process faces two main challenges: First, converting SNNs from low-bit ANNs can lead to ``occasional noise" --  the phenomenon where occasional spikes are generated in spiking neurons where they should not be -- during inference, which significantly lowers SNN accuracy. Second, although low-latency SNNs initially show fast improvements in accuracy with time steps, these accuracy growths soon plateau, resulting in their peak accuracy lagging behind both full-precision ANNs and traditional ``long-latency SNNs'' that prioritize precision over speed.

In response to these two challenges, this paper introduces a novel technique named ``noise adaptor.'' Noise adaptor can model occasional noise during training and implicitly optimize SNN accuracy, particularly at high simulation times $T$. Our research utilizes the ResNet model for a comprehensive analysis of the impact of the noise adaptor on low-latency SNNs. The results demonstrate that our method outperforms the previously reported  quant-ANN-to-SNN conversion technique. We achieved an accuracy of 95.95\%  within 4 time steps on CIFAR-10 using ResNet-18, and an accuracy of 74.37\% within 64 time steps on ImageNet using ResNet-50. Remarkably, these results were obtained without resorting to any noise correction methods during SNN inference, such as negative spikes or two-stage SNN simulations. Our approach significantly boosts the peak accuracy of low-latency SNNs, bringing them on par with the accuracy of full-precision ANNs. Code will be open source.

\end{abstract}

%% file: sec/1_intro.tex
\section{Introduction}
\label{sec:intro}

In recent years, deep spiking neural networks (SNNs) have gained substantial attention due to their rapid, efficient information processing abilities which closely mimic those of biological neural systems \cite{diehl2015fast,rueckauer2017conversion,li2023unleashing,deng2021optimal,wu2018spatio,zhang2018plasticity,meng2022training,fang2021deep,li2021free}. As a biologically plausible alternative to artificial neural networks (ANNs), SNNs provide valuable insights into the functioning of the human brain and have become one of the focal points for researchers in the pursuit of developing next-generation artificial intelligence (AI) systems \cite{maass1997networks}. This growing interest is driven by the potential benefits SNNs can offer in terms of power consumption, latency, and other performance metrics \cite{roy2019towards}, which promise significant advancements in the field of AI.

To fully harness the potential of SNNs, there have been rapid strides in both hardware and algorithms. On the hardware front, remarkable progress has been made in the evolution and enhancements of neuromorphic systems, including but not limited to SpiNNaker 2 \cite{mayr2019spinnaker}, BrainScaleS-2 \cite{pehle2022brainscales}, Loihi 2 \cite{orchard2021efficient}, and DYNAP-SE2 \cite{ccakal2023training}. These cutting-edge devices feature enhanced chip capacity, innovative solutions for efficient spike routing, and breakthroughs in architectural design. 

Concurrently, algorithmic research has demonstrated that SNNs can achieve accuracy levels on par with ANNs through ANN-to-SNN conversion \cite{diehl2015fast, rueckauer2017conversion, sengupta2019going}. However, this often requires long simulation times, negating the benefits of using SNNs in the first place. As a result, research has shifted towards to low-latency SNNs, also known as fast SNNs, which aim to achieve high inference accuracy with as few time steps as possible \cite{li2022quantization, li2023unleashing, hao2023bridging}. Given the heightened interest and fast progress in low-latency SNN algorithms, the latency required for SNNs to deliver satisfactory accuracy has consistently decreased from thousands of time steps \cite{rueckauer2017conversion, sengupta2019going} to hundreds \cite{deng2021optimal,li2021free} and, more recently, fewer than ten time steps \cite{li2021differentiable, fang2021deep}. 

Recently, a particularly interesting approach, quant-ANN-to-SNN conversion \cite{li2022quantization, bu2021optimal,meng2022training}, has been proposed. This method, which quantized the activations of ANN prior to their conversion into SNNs, yields promising outcomes in developing low-latency SNNs. Notably, it demands fewer memory and computational resources during training, which significantly accelerates the training process when compared to methods that rely on surrogate gradients \cite{neftci2019surrogate}. This makes quant-ANN-to-SNN conversion an attractive choice for researchers aiming to quickly develop low-latency SNNs.

The construction of low-latency SNNs through conversion from activation-quantized ANNs often encounters a significant challenge: the presence of occasional noise, also known as unevenness error \cite{li2022quantization, bu2021optimal,meng2022training}.  This issue stems from the variability in postsynaptic spike sequences that are not fully addressed during the training of low-bit ANNs. As a result, this occasional noise can lead to diminished accuracy in SNNs and increase the number of time steps required to achieve the desired accuracy level of the corresponding low-precision ANN. To mitigate this problem, various strategies have been proposed. These methods primarily focus on improving accuracy by incorporating additional artificial noise correction mechanisms during SNN inference \cite{li2022quantization,hu2023fast,hao2023bridging,hao2023reducing}. Despite these efforts, a comprehensive and effective solution for seamless, end-to-end conversion from low-bit ANNs to low-latency SNNs, while counteracting this challenge, remains elusive.

Another key limitation of quant-ANN-to-SNN conversion, identified for the first time in our paper, is the peak accuracy issue.  While quant-ANN-to-SNN conversion is effective in achieving satisfactory SNN accuracy within a limited number of time steps, it fails to guarantee a consistent increase in accuracy over time. We frequently observe a rapid plateau in accuracy growth, particularly noticeable in ultra-low-latency SNNs, resulting in a constrained peak accuracy for SNNs. This finding implies that the accelerated inference capabilities of SNNs, attained through the quant-ANN-to-SNN conversion, might be compromised by a reduction in overall accuracy. 

In this study, we introduce a novel approach called ``noise adaptor'' to mitigate both of these challenges in SNNs developed via quant-ANN-to-SNN conversion. Our contributions to this field are threefold:

\begin{enumerate}
\item \textbf{Tackling Occasional Noise}: We establish that the issue of occasional noise, a primary concern in current quant-ANN-to-SNN studies, can be effectively modelled and managed through a learning-based approach. Our research indicates that injecting controllable noise during training can better capture spike dynamics, thereby improving the tolerance to occasional noise.

\item \textbf{Enhancing Peak Accuracy}: We address the issue of low peak accuracy met by quant-ANN-to-SNN conversion. Our experiments show that noise adaptor significantly boosts the peak accuracy of low-latency SNNs, making it comparable to that of full-precision ANNs in some instances.

\item \textbf{Comprehensive Analysis Using ResNet}: Using ResNet as a case study, we conducted an in-depth analysis of the quant-ANN-to-SNN conversion process. This examination delves into various aspects such as the differences in pooling layers between ANNs and SNNs, as well as the effects of varying network depths.

\end{enumerate}

\begin{figure}
\centering
\includegraphics[width=0.465\textwidth]{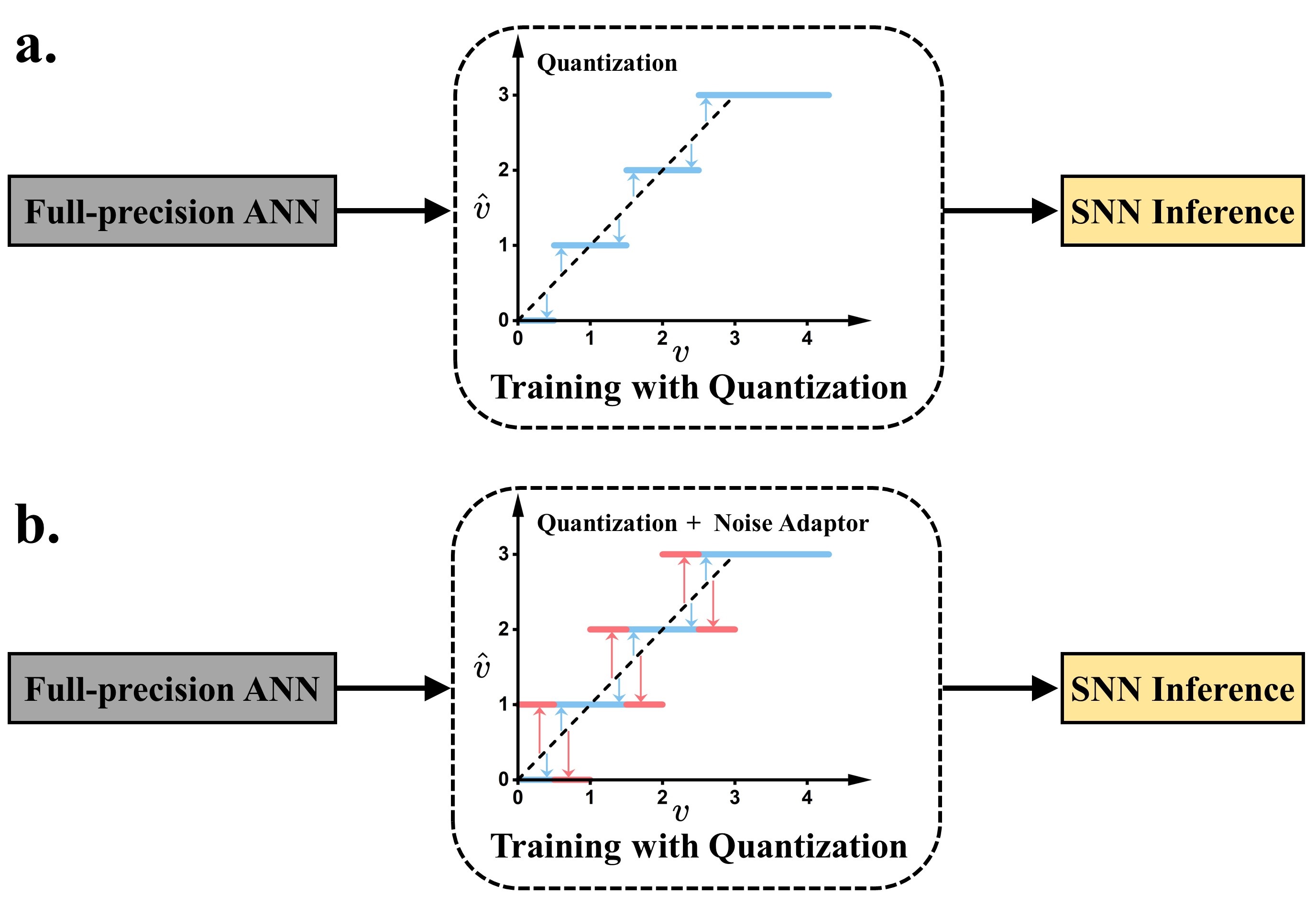} 

\caption{ \textbf{a.} In standard quant-ANN-to-SNN conversion, ANNs are trained with activation quantization to model the discrete spikes in SNNs. \textbf{b.} Our method incorporates random state transitions in the activation quantization function which alleviates the effects of occasional noise in low-latency SNNs.}
\label{fig1}
\end{figure}

%% file: sec/2_formatting.tex
\section{Preliminary}
\subsection{Spiking Neuronal Dynamics}
We use the integrate-and-fire model for spiking dynamics in this paper.  To allow for more precise information processing within spiking neurons, the reset mechanism of the neuronal model is changed from reset-to-resting-potential to reset-by-subtraction \cite{bu2021optimal,li2022quantization, rueckauer2017conversion,han2020rmp}. The equations governing this integrate-and-fire model are as follows:

\begin{equation}\label{eq1}
\bm{u}_{t}^{l} = \bm{u}_{t-\Delta{t}}^{l}+\widetilde{\bm{W}}_t^l\bm{z}_{t}^{l-1}{th}^{l-1}+\widetilde{\bm{B}}^l-\bm{z}_{t-\Delta{t}}^{l}{th}^{l}, 
\end{equation}

\begin{equation}\label{eq2}
\bm{z}_{t}^{l} =\Theta(\bm{u}_{t}^{l}-{th}^{l}).
\end{equation}

\noindent The first equation presented here primarily involves variables $\bm{u_}t^l$ and $\bm{u}_{t-\Delta{t}}^l$, which represent the membrane potential of spiking neurons situated in layer $l$ at two different instances: time $t$ and time $t-\Delta{t}$, respectively. Here, $\Delta{t}$ signifies the time resolution employed during the simulation. The variables $\widetilde{\bm{W}}_t^l$ and $\widetilde{\bm{B}}^l$ correspond to the synaptic weights and biases within layer $l$. The spikes generated by spiking neurons in layer $l$ at time $t$ are denoted by 
$\bm{z}_{t}^l$, which is 
calculated according to the membrane potential $\bm{u_}t^l$, the firing threshold ${th}^{l}$, and Heaviside step function $\Theta(\cdot)$, as shown in the second equation.

In deep SNNs, the synaptic models employed are often simpler than those used in traditional SNN simulations \cite{bogdan2021towards,gammaitoni1998stochastic}. This simplification facilitates more efficient simulation on GPUs. In these models, complex synaptic dynamics are typically not considered. Instead, synaptic weights are treated as constant and time-invariant. This concept is encapsulated in the equation  $\widetilde{\bm{W}}_t^l= \widetilde{\bm{W}}^l$, where \(\widetilde{\bm{W}}_t^l\) represents the synaptic weight at any given time \(t\), and \(\widetilde{\bm{W}}^l\) denotes the constant synaptic weight.

\subsection{Forward and Backward Pass of ANN Activation Quantization}

In the standard ANN quantization technique, the quantization operator applied to activation, with ReLU as a representative example, can be formalized as follows:

\begin{equation}\label{eq3}
	\hat{v}^l = s^l\left\lfloor clip \left( \frac{v^l}{s^l}, 0, p\right) \right\rceil.
\end{equation}

\noindent In this equation, \(v^l\) represents the original, full-precision value in layer \(l\) before quantization, while \(\hat{v}^l\) denotes the quantized activation value after the quantization process. The term $s^l$ is the quantization scale factor. The function $clip(z, 0, p)$ truncates all elements within $z$ to the range of $[0, p]$, where $p$ signifies the positive integer threshold of the activation\footnote{In ANN activation quantization, $p$
is typically derived using the formula $p=2^b-1$, with $b$ being the bit precision. Nevertheless, when it comes to quant-ANN-to-SNN conversion,  the value of $p$ is more flexible and it can be any positive integer. 
}. The rounding operation $\lfloor z \rceil$  rounds each element in $z$ to the nearest integer.

During the backward phase of ANN quantization training, the non-differentiability of \(\frac{\partial \hat{v}^l}{\partial v^l}\) is addressed by applying the straight-through estimator \cite{bengio2013estimating}. The gradients are then given by

\begin{equation}\label{eq4}
\frac{\partial{\hat{v}^l}}{\partial{v^l}} =
\begin{cases}
0								& \text{if $\frac{v^l}{s^l} \leqslant 0$} \\
1			& \text{if $0 < \frac{v^l}{s^l} < p$} \\
0.							& \text{if $\frac{v^l}{s^l} \geqslant p$} \\
\end{cases}
\end{equation}

\noindent It has been observed that allowing the quantization scale factor \(s^l\) to be learnable can lead to a more optimal local minimum and improve quantization accuracy. Following the strong baseline quantization method LSQ \cite{esser2019learned}, the gradient is defined as 

\begin{equation}\label{eq5}
\frac{\partial{\hat{v}^l}}{\partial{s^l}} =
\begin{cases}
0									& \text{if $\frac{v^l}{s^l} \leqslant 0$} \\
-\frac{v^l}{s^l} + \lfloor \frac{v^l}{s^l} \rceil			& \text{if $0 < \frac{v^l}{s^l} < p$} \\
p.									& \text{if $\frac{v^l}{s^l} \geqslant p$} \\
\end{cases}
\end{equation}

\subsection{Quant-ANN-to-SNN conversion}

The process of converting an artificial neural network into a spiking neural network is referred to as ANN-to-SNN conversion \cite{diehl2015fast,rueckauer2017conversion}.  This conversion process hinges on the principle of harnessing the input-output response curve of spiking neurons to emulate the activation function (such as ReLU and quantized ReLU) used in ANNs.

However, while ANN neurons operate through continuous activation functions, their SNN counterparts communicate via discrete events known as spikes, which occur at specific moments. This significant difference has led to the development of several algorithms designed to further optimize the ANN-to-SNN conversion process such as the popular   quant-ANN-to-SNN conversion algorithm \cite{li2022quantization, bu2021optimal,meng2022training}. This technique quantises the ANN activation prior to the actual conversion taking place. Quantization serves as a means of compressing information, thus decreasing the number of spikes required by the spiking neurons to represent ANN activations \cite{li2022quantization}. This technique has been widely reported to bring considerable savings in spike counts and SNN inference latency \cite{li2022quantization, bu2021optimal,meng2022training, hu2023fast,hao2023bridging,hao2023reducing}. Moreover, the quantization operation and the clipping point used in the quantized ANN activation effectively emulate two essential characteristics of SNNs: the discreteness of spikes and the cut-off frequency (the upper limit firing rate for spiking neurons). This approach helps to minimize the accuracy drop incurred during the ANN-to-SNN conversion.

Upon reviewing Equation \ref{eq1} and Equation \ref{eq2}, which define dynamics of spiking neurons, we identify three parameters in SNNs—$\widetilde{\bm{W}}^l$, $\widetilde{\bm{B}}^l$, and ${th}^{l}$—that require computation from ANN parameters during quant-ANN-to-SNN conversion. The corresponding conversion equations are outlined as follows:

\begin{equation}\label{eq6}
\left\{
\begin{aligned}
   \widetilde{\bm{W}}^1 &= \bm{W}^l, \\
   \widetilde{\bm{B}}^1 &= \bm{B}^l, \\
   th^l &= ps^l.
\end{aligned}
\right.
\end{equation}

\noindent Here, $\bm{W}^l$ and $\bm{B}^l$ symbolize the weight and bias of layer $l$ in the ANN, respectively. The spiking threshold ${th}^l$ is a function of the positive integer threshold $q$ and the quantization step size ${s^l}$ as defined in ANN activation quantization (refer to Equation \ref{eq3}).

The integrate-and-fire mechanism in SNNs is analogous to the process of ``flooring'' or rounding down in ANNs, in contrast to the ``round to nearest'' method used in ANN quantization. The prevailing quant-ANN-to-SNN conversion algorithms compensate for this discrepancy by initializing the membrane potential at \(0.5 \text{th}^l\) across all hidden layers at the start of SNN simulation \cite{li2022quantization, bu2021optimal}.

%% file: sec/3_finalcopy.tex
\section{Related Work}
\paragraph{Fast SNNs.}
As the demand for real-time, energy-efficient computing grows, fast SNNs have come to the forefront of contemporary SNN research. Our work primarily focuses on quant-ANN-to-SNN conversion \cite{li2022quantization, bu2021optimal,meng2022training}, one of the leading techniques for developing fast SNNs, along with surrogate gradients \cite{neftci2019surrogate,shrestha2018slayer,wu2018spatio,fang2021deep}. It's widely acknowledged in numerous published studies that, when an ANN is trained with quantized activations to mimic discrete spike counts, it becomes more amenable to conversion into an SNN compared to those trained without such constraints \cite{li2022quantization, bu2021optimal,meng2022training}. A notable challenge that quant-ANN-to-SNN conversion meets is that the reduction in latency amplifies the impact of occasional noise \cite{li2022quantization}, which substantially impedes the achievement of high accuracy within limited time steps. QCFS \cite{bu2021optimal} provides a theoretical analysis of the occasional noise and amortizes this noise by extending the simulation duration. QFFS \cite{li2022quantization} identifies occasional noise at the runtime of SNN and generates negative spikes to correct noise. SRP \cite{hao2023reducing} and COS \cite{hao2023bridging} utilize a two-stage approach for noise reduction: initially running an SNN for multiple time steps to identify occasional noise followed by a second run to correct it. In summary, existing methods for occasional noise suppression rely on either increasing SNN response times or adopting artificial interventions during SNN runtime to identify and amend the occasional noise. Consequently, finding an efficient method to minimize the impact of occasional noise in SNNs remains a substantial and unresolved challenge.

\paragraph{Noise in SNNs.}
A fundamental trait of SNNs is the existence of noise \cite{kempter1998extracting,stein2005neuronal,faisal2008noise,ma2023exploiting}. Understanding the role and implications of noise is crucial for improving the robustness and functionality of SNNs. During the process of ANN-to-SNN conversion, SNNs are subject to multiple noise sources that can impair inference performance, such as rate-coding noise, sub-threshold noise, and supra-threshold noise \cite{diehl2015fast}. To counteract these effects, a range of strategies have been developed, including the use of analog input coding \cite{rueckauer2017conversion}, soft-reset (or reset-by-subtraction) \cite{han2020rmp,rueckauer2017conversion}, bias shift \cite{deng2021optimal}, and pre-charging the membrane potential \cite{hwang2021low}. For applications aiming to create low-latency or ultra-low-latency SNNs by ANN-to-SNN conversion, it is necessary to address and minimize occasional noise to achieve the desired accuracy \cite{li2022quantization}. Moreover, when deploying SNNs on neuromorphic hardware \cite{schemmel2010wafer}, especially on those utilizing noisy components like memristors \cite{burr2017neuromorphic} and skyrmions \cite{chen2020nanoscale} for weight storage, it is critical to account for noise. Doing so helps prevent system malfunctions and significant accuracy losses, ensuring the optimal functioning of spike-based computation systems.

%% file: sec/4.tex
\section{Methodology}
\subsection{Rethinking Occasional Noise Management}

The quant-ANN-to-SNN conversion process not only defines how the SNN parameters, such as weights, biases, and firing thresholds, are calculated through ANN training but also determines how occasional noise is managed. Existing studies on quant-ANN-to-SNN conversion  defer handling of occasional noise until the SNN inference phase,  due to our limited understanding of its detailed dynamics. The effect of this noise varies across different samples, layers, and spiking neurons, making its impact unpredictable until the actual running of the SNN. Furthermore, the pattern of occasional noise cannot be estimated through a small calibration set, as occasional noise impacts the inference results of different input samples in unique ways. This inherent unpredictability of the dynamics of occasional noise poses significant challenges in addressing it prior to running SNNs.  Consequently, current research mainly focuses on mitigating the negative impact of occasional noise when it becomes measurable during SNN runtime \cite{li2022quantization,hao2023reducing,hao2023bridging}.

\subsection{Noise Adaptor}
Our research proposes a novel approach that manages occasional noise during the ANN training phase. This is essential for further mitigating the discrepancies between the simplified neuron models used in ANNs in training and the complex neuron dynamics observed in SNNs during inference.

\noindent\textbf{Forward Pass}: The noise adaptor is a flexible, plug-and-play module that can fit smoothly into the ANN activation quantization process. Its primary function is to introduce a controlled noise component to the quantization operator, as described below:

\begin{equation}\label{eq7}
	\hat{v}^l = s^l\left\lfloor clip\left( \frac{v^l}{s^l}+\epsilon, 0, p\right) \right\rceil,
\end{equation}

\begin{equation}\label{eq8}
    \epsilon = U(-0.5, 0.5),
\end{equation}

\noindent where $\epsilon$ represents a uniformly distributed noise term over the interval $(-0.5, 0.5)$. The dimensions of this noise term are identical to those of $v^l$. During each forward pass, $\epsilon$ is sampled. The sampled $\epsilon$ is stored and then used in the subsequent backward pass. This approach ensures that the noise influence remains consistent throughout both forward pass and backward pass.

\noindent\textbf{Backward Pass}: During the backward phase of ANN quantization training, the gradients $\frac{\partial{\hat{v}^l}}{\partial{v^l}}$ and $\frac{\partial{\hat{v}^l}}{\partial{s^l}}$ are calculated as follows:

\begin{equation}\label{eq9}
\frac{\partial{\hat{v}^l}}{\partial{v^l}} =
\begin{cases}
0								& \text{if $\frac{v^l}{s^l}+\epsilon \leqslant 0$} \\
1			& \text{if $0 < \frac{v^l}{s^l}+\epsilon  < p$} \\
0,							& \text{if $\frac{v^l}{s^l}+\epsilon  \geqslant p$} \\
\end{cases}
\end{equation}

\begin{equation}\label{eq10}
\frac{\partial{\hat{v}^l}}{\partial{s^l}} =
\begin{cases}
0									& \text{if $\frac{v^l}{s^l}+\epsilon \leqslant 0$} \\
-(\frac{v^l}{s^l}+\epsilon) + \lfloor \frac{v^l}{s^l} +\epsilon\rceil			& \text{if $0 < \frac{v^l}{s^l}+\epsilon < p$} \\
p.									& \text{if $\frac{v^l}{s^l}+\epsilon \geqslant p$} \\
\end{cases}
\end{equation}

 \noindent\textbf{Comprehensive Understanding}: Figure \ref{fig1} illustrates the effect of a noise adaptor on the activation quantization function in quant-ANN-to-SNN conversion (with $p=3$, $s=1$). As shown in Figure \ref{fig1}.a, standard conversion processes involve a fixed direction for rounding each variable $v$, either upwards or downwards. However, as depicted in Figure \ref{fig1}.b, the introduction of a noise parameter $\epsilon$ introduces an additional possibility, allowing rounding to an alternative state, thereby adding new dynamics into activation.

\noindent\textbf{Conversion Process}: The proposed noise adaptor does not alter the core process of converting a quantized ANN to an SNN model. After ANN training, the model can be converted to an SNN model using Equation \ref{eq6}, accompanied by initializing the membrane potential to \(0.5 {th}^l\).

\subsection{Related Considerations}

\noindent\textbf{Stochastic Quantization Outputs}: After integrating a noise term into the activation quantization function of ANNs, the activation output $\hat{v}^l$ becomes stochastic and capable of transitioning to adjacent discrete states either upwards or downwards by $s^l$. This stochasticity mirrors the behavior of spiking neurons, which, under the influence of occasional noise, can either increase or decrease their spike count, typically by one \cite{hao2023reducing}.

\noindent\textbf{Controlling Noise Amplitude}: It is also reported that the likelihood of spiking neurons erroneously generating more or fewer than one spike increases with the complexity of the dataset \cite{hao2023bridging}. However, accurately modeling this phenomenon is challenging: To simulate larger spike jumps, it is necessary to inject noise with a higher amplitude, which can destabilize the training process, especially with intricate datasets like ImageNet. Therefore, our approach only models the spike count change of  one, up or down. We achieve this by restricting the noise range within $(-0.5, 0.5)$, as outlined in Equation \ref{eq8}.

\noindent\textbf{Probability of State Transitions}: We propose that the likelihood of state transitions in a spiking neuron should be related to its input values. For instance, a spiking neuron with an input of $0.99 \text{th}^l$ is more likely to produce an additional spike in response to occasional noise than one with an input of $0.01 \text{th}^l$. The closer the input is to the threshold of the next quantization level, the higher the chance of transitioning to a different integer state. To model this effect, we add the noise term prior to the clipping and rounding steps in the ANN activation function. This allows the ratio $\frac{v^l}{s^l}$ to influence the likelihood of state changes.

\noindent\textbf{Impact  on Backward Gradients}: The noise term $\epsilon$ also affects the backward gradients. Generally, it introduces additional noise into these gradients\footnote{Typically, introducing unbiased noise with a zero mean into the gradients can enhance training effectiveness, provided that its amplitude remains constrained.}. This extra noise facilitates exploring a broader range of the loss landscapes, potentially leading to the discovery of better local minima.

\subsection{The ``Second'' Chance of Effective Conversion}

Current theoretical research on the topic of quant-ANN-to-SNN conversion primarily focuses on minimizing the disparity between the response curve of spiking neurons and the quantized activations in ANN. A pivotal aspect of this research is the alignment of the SNN simulation time \(T\) with the number of positive activation states \(p\) employed during the quantization training of the ANN. This alignment, however, poses a challenge due to the influence of T on the spiking neurons' response curve in SNN, complicating the precise matching with the quantized ANN activation.

Our study introduces a novel perspective, termed the ``second chance'' of effective conversion in quant-ANN-to-SNN conversion. We observe that with the increase in \(T\), the response curve of spiking neurons more closely aligns with the rectified-ReLU function. This alignment becomes more pronounced at higher $T$ values, where the impact of occasional noise diminishes. Leveraging this, we propose a dual-optimization target approach during ANN quantization training. The first target adheres to the traditional ANN activation quantization methods, focusing on the learnable parameters $\bm{W}^l$, $\bm{B}^l$, and \(s^l\). The second target involves optimizing the ANN with an activation function that mimics rectified-ReLU, determined by a rectified point set at \(ps^l\). This dual-target optimization aims to refine the response curve of spiking neurons for both \(T=p\) and \(T=\infty\), ensuring SNN accuracy improves over simulation time step $T$ in [p, $\infty$].

Incorporating noise adaptor in the ANN quantization process can benefit to achieving the second goal mentioned above. The noise adaptor ensures the alignment of the expected mean of \({s^l}\left\lfloor \frac{v^l}{s^l} + \epsilon \right\rceil\) with the actual value \(v^l\). This approach effectively transforms Equation \ref{eq7} into a variant that resembles a rectified-ReLU function, represented as \(\hat{v}^l = \text{clip}(v^l, 0, s^{l}p)\). While this modification allows for implicit optimization of the ANN using this activation function during training, it is important to note that it also introduces a certain level of jitter in the activation process.

This approach potentially revolutionizes quant-ANN-to-SNN conversion. By leveraging the rectified-ReLU function, which generally achieves accuracy comparable to the standard ReLU, our method suggests that low-latency SNNs built by quant-ANN-to-SNN conversion can attain similar accuracy levels to full-precision ANNs, especially at longer simulation time steps \(T\).

%% file: sec/5.tex
\section{Experimental Results}

\subsection{Experimental Setup}

We validate the effectiveness of our method on the CIFAR-10 and ImageNet image classification datasets using the ResNet network architecture. We conduct ANN quantization training based on pre-trained full-precision ResNet models. Before training, we replace max-pooling with average-pooling and substitute the ReLU activation with the quantization function as described in Equation~\ref{eq7}. The network weights are maintained at full precision. Training is performed using stochastic gradient descent, employing a cross-entropy loss function and a cosine learning rate scheduler. The initial learning rate is set to $0.1$, with a weight decay of $5 \times 10^{-4}$ for CIFAR-10 and $2.5 \times 10^{-5}$ for ImageNet. Each model is trained for 160 epochs, selecting $p=2$ for CIFAR-10 and $p=3$ for ImageNet, unless otherwise specified. 

After training, we transfer all weights and biases from the source ANN to the converted SNN, setting the threshold $th^l$ as per Equation~\ref{eq6}. To mitigate systematic errors associated with the round-to-nearest method used in ANN quantization, the membrane potential of the spiking neurons in all hidden layers is pre-charged to $0.5{th}^l$ at the initial time step. In the SNN simulation, inputs are analog-coded \cite{rueckauer2017conversion}, with a time resolution $\Delta{t}$ of $1\,$ms. The adopted neuronal model is described in Equations~\ref{eq1} and \ref{eq2}.

\subsection{Effectiveness }

In Table \ref{tab1}, we assess the impact of incorporating the proposed noise adaptor in the quant-ANN-to-SNN conversion process. Our results indicate that using a noise adaptor considerably boosts SNN accuracy at each time step. This implies that noise adaptor not only enables SNNs to achieve higher accuracy levels swiftly during the initial time steps but also supports their continued accuracy improvement over time. However, we observed a decrease in ANN accuracy when the noise adaptor was used. This reduction in ANN performance could be attributed to the introduction of a stochastic activation function by the noise adaptor, potentially impacting ANN accuracy negatively.

Interestingly, our findings also highlight that high ANN accuracy to begin with is not essential for attaining a high SNN accuracy. In experiments with CIFAR-10 and ImageNet, the models without a noise adaptor outperformed those with it in terms of ANN accuracy, by margins of 0.53\% and 1.01\%, respectively. Despite this, the inclusion of the noise adaptor consistently yielded superior SNN inference accuracy at all time steps. Specifically, on the CIFAR-10 dataset, models with a noise adaptor required half the time steps to reach the same or greater accuracy compared to those without it. These results indicate that the key element for successful conversion from ANN to SNN and achieving optimal SNN performance is the degree of similarity between the ANN and SNN models. This insight underpins our proposal of the noise adaptor, aiming to minimize this discrepancy during the training phase.

\begin{table}[htb!]
    \centering

     \resizebox{\columnwidth}{!}
     {
    \begin{tabular}{ccccccc}
    \toprule
    \multirow{2}{*}{\textbf{CIFAR-10 Setting}} & \multirow{2}{*}{\textbf{ANN}} & \multicolumn{5}{c}{$\pmb{T}$}\\
    \cmidrule(lr){3-7}

   & &T=1 & T=2 & T=4 & T=8 & T=16  \\
    \midrule
    
ResNet-18 w/o NA	&\textbf{95.50}&91.62& 93.65 &95.18& 96.08 & 96.38 \\
ResNet-18 w/ NA  &94.97&\textbf{93.71}& \textbf{95.26} &\textbf{95.95}& \textbf{96.61} & \textbf{96.72}	\\

\midrule[\heavyrulewidth]

    \multirow{2}{*}{\textbf{ImageNet Setting}} & \multirow{2}{*}{\textbf{ANN}} & \multicolumn{5}{c}{$\pmb{T}$}\\
    \cmidrule(lr){3-7}

   & & T=8 & T=16 & T=32 & T=64 & T=128  \\
    \midrule
    
ResNet-50 w/o NA&\textbf{74.73}&21.19&48.59& 67.47 &73.47 & 74.64 \\
ResNet-50 w/ NA  &73.72&\textbf{36.00}&\textbf{59.05}&\textbf{70.82} &\textbf{74.37}& \textbf{75.07}	\\

      \bottomrule
    \end{tabular} 
     } 
    \vspace{-1mm}
    \caption{Comparisons of quant-ANN-to-SNN conversion with and without noise adaptor (NA). We benchmark ANN accuracy after training with activation quantization, as well as SNN accuracy at different time steps $T$.}
    \vspace{-1mm}
    \label{tab1}
\end{table}

\begin{table}[htb!]
    \centering

     \resizebox{\columnwidth}{!}
     {
    \begin{tabular}{cccc}
    \toprule

    \textbf{Network Architecture} & \textbf{Training Epochs} & \textbf{ANN Accuracy} & \textbf{SNN Accuracy (T=128)}\\

    \midrule
    
ResNet-50 &160&73.72\%&75.07\% \\
ResNet-50 &90&72.40\%&73.49\%	\\
ResNet-50 &40&70.72\%&71.97\%	\\
\midrule
ResNet-18 &40&67.07\%&68.85\%\\
ResNet-34 &40&69.72\%& 71.35\%	\\
ResNet-50 &40&70.72\%&71.97\%	\\
ResNet-101 &40&72.92\%&73.19\%	\\

      \bottomrule
    \end{tabular} 
     } 
    \vspace{-1mm}
    \caption{Influence of training epochs and network depth. The dataset is ImageNet. }
    \vspace{-1mm}
    \label{tab2}
\end{table}

\begin{table*}[tb]
\centering
\footnotesize
\resizebox{\textwidth}{!}
    {
    \begin{tabular}{cccccccccccccc}
    \toprule
     \textbf{Methods}&\textbf{Datasets}&  \textbf{Architecture}& \textbf{$p$}& \textbf{ANN}  & \textbf{T=1} & \textbf{T=2} & \textbf{T=4} & \textbf{T=8} & \textbf{T=16} & \textbf{T=32} & \textbf{T=64} & \textbf{T=128}& \textbf{T} $\geqslant$\textbf{256}\\
    \midrule

    QCFS\cite{bu2021optimal} & \multirow{8}{*}{CIFAR-10} & \multirow{8}{*}{ResNet-18} & 4 & 96.04\% & - & 75.44 & 90.43\% & 94.82\% & 95.92\%& 96.08\% &96.06\% & - &96.06\%\\
    SlipReLU~\cite{jiang2023unified} &  &  & 1 & 94.61\% & 93.11\% & 93.97\% & 94.59\% & 94.92\% & 95.18\% & 95.07\% & 94.81\% & - &94.67\%\\
    SlipReLU~\cite{jiang2023unified} &  &  & 4 & \textbf{96.15\%}
    & 86.52\% & 90.78\% & 93.84\% & 95.48\% & 96.10\% & 96.12\% & 96.22\% & - & 96.15\%\\

    TTRBR~\cite{meng2022training} &  &  & 32 & 95.27\% & - & - & - & -& 93.99\% & 94.77\% & 95.04\% & 95.18\% & -\\
  
    \textbf{NoiseAdaptor} &                     &  & 1 & 93.65\%&\textbf{94.38\%}& 94.96\% &95.35\%& 95.57\% & 95.69\% &95.69\% &95.68\% &95.63\% &95.66\%\\

    \textbf{NoiseAdaptor} &                     &  & 2 &94.97\%&93.71\%& \textbf{95.26\%} &\textbf{95.95\%}& \textbf{96.61\%} & \textbf{96.72\%} &\textbf{96.74\%} &\textbf{96.72\%} &\textbf{96.69\%} &\textbf{96.68\%}\\

    \textbf{NoiseAdaptor} &                     &  & 3 &96.00\%&90.60\%& 93.35\% &95.08\%& 96.12\% & 96.52\% &96.44\% &96.47\% &96.46\%&96.44\%\\

    \textbf{NoiseAdaptor} &                     &  & 4 &95.98\%&87.11\%& 91.40\% &93.98\%& 95.67\% & 96.19\%  &96.30\% &96.29\% &96.25\%&96.24\%\\

    \midrule

    QCFS\cite{bu2021optimal} & \multirow{7}{*}{ImageNet} & ResNet-34 & 8 & 74.32\% & - & - & - & - & \textbf{59.35\%}& 69.37\% &72.35\% & 73.15\% &73.39\%\\
    
    SlipReLU~\cite{jiang2023unified} &  & ResNet-34 & 8 & 75.08\% & -& - & - & - & 43.76\% & 66.61\% & 72.71\% & 74.01\% &-\\

    TTRBR~\cite{meng2022training} &  & ResNet-34 & 32 & 74.24\% & - & - & - & -& - & - & - & - & 74.18\%\\
    
    TTRBR~\cite{meng2022training} &  & ResNet-50 & 32 & 76.02\% & - & - & - & -& - & - & - & - & 75.04\%\\

    TTRBR~\cite{meng2022training} &  & ResNet-101 & 32 & \textbf{76.82}\% & - & - & - & -& - & - & - & - & 75.72\%\\

    \textbf{NoiseAdaptor} &                    & ResNet-50 & 3 &73.72\%&7.79\%& 9.97\% &17.00\%& 35.99\% 
    & 59.05\%  &\textbf{70.82\%} &\textbf{74.37\%} &75.07\%&75.24\%\\

    \textbf{NoiseAdaptor} &                    & ResNet-50 & 7 &75.95\%&0.76\%& 0.8\% &1.08\%& 2.44\% 
    & 13.51\%  &49.56\% &70.46\% &\textbf{75.25\%}&\textbf{76.48\%}\\

    \bottomrule
    \end{tabular}}
    \vspace{0.2mm}
        \caption{Comparisons with other quant-ANN-to-SNN conversion algorithms. The comparisons are limited within algorithms without noise correction during SNN inference. $p$ is the positive integer threshold during ANN training.}
        \vspace{-0.2cm}
    \label{tab3}
\end{table*}

\subsection{Efficiency of Training}

In this section, we explore methods to lower the training costs associated with quant-ANN-to-SNN conversion. Traditionally, this conversion requires extensive training, typically up to 300 epochs from scratch, as indicated in the prior study \cite{bu2021optimal}. We introduce a more efficient alternative to this training practice.

A key challenge in quant-ANN-to-SNN conversion has been the disparity in pooling layers: SNNs typically employ average-pooling, whereas ANNs prefer max-pooling. This difference has traditionally necessitated training from scratch. However, our research challenges this notion. In our experiments, we start with pre-trained ANN models employing max-pooling and adopt them for ANN training with activation quantization. Our results, as presented in Table \ref{tab1}, show that despite the differences in pooling layers, competitive accuracy can be achievable after training. This discovery undermines the assumed criticality of pooling methods in the context of ANN-to-SNN conversion. By utilizing pre-trained, full-precision models, we have effectively reduced the training epochs from 300 to just 160.

This observed robustness to different pooling layers also presents opportunities for transfer learning. For instance, to enhance SNN accuracy on CIFAR-10, one could start with a model pre-trained on ImageNet and refine it through quant-ANN-to-SNN conversion, irrespective of the pooling method used. Similarly, for enhancing SNN accuracy on ImageNet, it is feasible to utilize ANN models trained on larger datasets, such as JFT-300M and ImageNet-21k.

Additionally, we examined the impact of reducing training epochs further, to 90 and 40, as detailed in Table \ref{tab2}. The findings suggest that while fewer training epochs may slightly decrease ANN accuracy, they do not significantly affect the effectiveness of the noise adaptor. Thus, the ANN-to-SNN conversion process remains efficient even with reduced training epochs, without a substantial loss in conversion quality.

\subsection{Comparision to Other Quant-ANN-to-SNN Conversion Methods}

This section compares our proposed noise adaptor method with established quant-ANN-to-SNN conversion techniques, specifically on the CIFAR-10 and ImageNet datasets. We focus on contrasting our approach with methods such as QCFS \cite{ding2021optimal}, SlipReLU \cite{jiang2023unified}, and TTRBR \cite{meng2022training}. Notably, these techniques do not incorporate noise correction strategies like negative spikes or two-stage simulations in SNN inference. Details on how our noise adaptor complements these noise correction techniques are discussed in the subsequent section.

Our noise adaptor method, when applied to a ResNet-18 architecture on the CIFAR-10 dataset, demonstrates exceptional accuracy improvements at various time steps. As shown in Table \ref{tab3}, with a parameter setting of $p=2$, our method attains accuracies of 95.26\% and 95.95\% at 
$T=2$ and $T=4$ respectively. This performance exceeds that of QCFS, SlipReLU, and TTRBR by a margin of at least 1.2\%. This consistent superiority across different time steps highlights the effectiveness of our noise adaptor in building competitive low-latency SNNs.

The performance of our method extends impressively to the more complex ImageNet dataset. Here, the noise adaptor demonstrates its distinct performance advantage, especially at longer time steps surpassing $T=32$. When $p=7$, our method achieves a remarkable peak accuracy of 76.48\%, significantly outperforming other results. This performance is on par with that of full-precision accuracy models. This underscores the adaptability and efficiency of our noise adaptor in managing more complex and larger datasets.

\subsection{Compatibility with Other Error Correction Techniques}

It is crucial to acknowledge that the methods we have discussed—QCFS, SlipReLU, and TTRBR—do not incorporate noise correction during SNN inference. This deliberate choice stems from our aim to address challenges during the training phase while maintaining a simple and cost-effective inference process. However, when noise correction techniques are combined with noise adaptor during SNN inference, it's possible to achieve enhanced performance.

Our experiments, detailed in Table \ref{tab4} and \ref{tab5}, explore the compatibility of the noise adaptor with various noise correction methods including negative spikes \cite{li2022quantization} and 2-stage offset spikes correction \cite{hao2023bridging}, using the CIFAR-10 and ImageNet datasets. The results, especially on CIFAR-10, are particularly noteworthy. They show that the noise adaptor alone can attain impressively high accuracy, making the integration of an additional noise correction mechanism potentially unnecessary. This finding is significant for researchers focusing on developing low-latency SNNs for compact edge-computing applications.

\begin{table}[htb!]
    \centering

     \resizebox{\columnwidth}{!}
     {
    \begin{tabular}{ccccccccc}
    \toprule
    \multirow{2}{*}{\textbf{Noise Correction Method}} & \multirow{2}{*}{\textbf{ANN}} & \multicolumn{5}{c}{$\pmb{T}$}\\
    \cmidrule(lr){3-8}

   & &T=1 & T=2 & T=4 & T=8 & T=16 & T $\geqslant$ 256  \\
    \midrule
None	&94.97&93.71& 95.26 &95.95& 96.61 & 96.72 & 96.68  \\
Negative Spikes	&94.97&93.71& 95.87 &96.55& 96.70 & 96.71& 96.73 \\

Offset Spikes (lightweight)	&94.97&95.82& 96.25 &96.33& 96.66 & 96.67& 96.66 \\
      \bottomrule
    \end{tabular} 
     } 
    \vspace{-1mm}
    \caption{The compatibility of noise adaptor with noise corrections techniques applied during SNN inference. The network architecture is ResNet-18 and the dataset is CIFAR-10. Accuracy is expressed as percentage (\%).}
    \vspace{-1mm}
    \label{tab4}
\end{table}

\begin{table}[htb!]
    \centering

     \resizebox{\columnwidth}{!}
     {
    \begin{tabular}{ccccccccc}
    \toprule
    \multirow{2}{*}{\textbf{Noise Correction Method}} & \multirow{2}{*}{\textbf{ANN}} & \multicolumn{5}{c}{$\pmb{T}$}\\
    \cmidrule(lr){3-8}

   & &T=1 & T=2 & T=4 & T=8 & T=16 & T $\geqslant$ 256  \\
    \midrule
None	&73.72&7.79& 9.97 &17.00& 35.99 & 59.05 & 75.24  \\
Negative Spikes	&73.72&7.79& 53.42 &69.74& 72.88 & 73.94& 74.72 \\
Offset Spikes (lightweight)	&73.72&63.42& 64.84 &66.34& 68.43 & 70.60& 73.37 \\
      \bottomrule
    \end{tabular} 
     } 
    \vspace{-1mm}
    \caption{The compatibility of noise adaptor with noise corrections techniques that applied during SNN inference. The network architecture is ResNet-50 and the dataset is ImageNet. Accuracy is expressed as percentage (\%).}
    \vspace{-1mm}
    \label{tab5}
\end{table}

\subsection{Impact of Network Depth}

In the lower section of Table \ref{tab2}, we examine the performance of the noise adaptor across various ResNet architectures implemented on ImageNet. A key observation is that as network depth increases, there is a corresponding improvement in the accuracy of both ANN and SNN. Notably, there is no evidence of failure in the noise adaptor during the ANN-to-SNN conversion process, even in networks with up to 101 layers. This consistency across different network depths indicates that the noise adaptor's effectiveness is not compromised by the complexity of the network architecture. This highlights the robustness and reliability of the noise adaptor in facilitating smooth ANN-to-SNN conversions across a range of network depths.

\subsection{Impact of the choice of $p$}

In the context of low-bit ANN training and its application in quant-ANN-to-SNN conversion, the positive integer threshold \( p \) plays a pivotal role as a hyperparameter. Table \ref{tab3} illustrates that a higher \( p \) generally enhances ANN accuracy. However, it is noteworthy that for a corresponding SNN, achieving this level of accuracy necessitates an increased number of time steps.

Our studies on the CIFAR-10 dataset highlight specific optimal scenarios or ``sweet spots''. For instance, at \( p=1 \), the model shows a remarkable initial accuracy of 94.38\% in the first time step, with a continuous improvement over time. Despite this, it is important to note that for longer durations (higher \( T \) values), its performance is outmatched by other \( p \) settings. On the other hand, setting \( p \) to 2 offers the best overall accuracy across various time steps \( T \), except for the very first step (\( T=1 \)).

%% file: sec/7.tex
\section{Conclusion}

This study introduced the ``noise adaptor," a novel technique to improve the accuracy of low-latency spiking neural networks converted from low-bit artificial neural networks. This approach effectively addresses two main challenges: the occurrence of occasional noise during inference and the plateauing of accuracy improvements in low-latency SNNs. Our implementation using the ResNet model demonstrated significant results, achieving 95.95\% accuracy on CIFAR-10 with ResNet-18 and 74.37\% on ImageNet with ResNet-50, without any additional noise correction methods during SNN inference.

The noise adaptor method surpasses previous quant-ANN-to-SNN conversion techniques, bringing the accuracy of low-latency SNNs in line with full-precision ANNs. This advancement not only solves key issues in SNN accuracy but also opens up new avenues for their application in fast and accurate computing scenarios.